\documentclass{Interspeech2024}

\newcommand{\orange}[1]{\textcolor{orange}{#1}}

\usepackage{tcolorbox}
\usepackage{lipsum}
\newcommand\blfootnote[1]{%
  \begingroup
  \renewcommand\thefootnote{}\footnote{#1}%
  \addtocounter{footnote}{-1}%
  \endgroup
}

\usepackage{amsmath,amsfonts}
\usepackage{microtype}
\usepackage{graphicx}
\usepackage{subcaption}
\usepackage{booktabs}
\usepackage{pgfplots,pgfplotstable}
\pgfplotsset{compat=1.17}
\usepackage{multirow}

\usepackage{hyperref}
\usepackage{url}

\usepackage[capitalize,noabbrev]{cleveref}

\definecolor{forestgreen}{rgb}{0.13, 0.55, 0.13}
\definecolor{fulvous}{rgb}{0.86, 0.52, 0.0}
\definecolor{glaucous}{rgb}{0.38, 0.51, 0.71}
\definecolor{lava}{rgb}{0.81, 0.06, 0.13}
\definecolor{buff}{rgb}{0.94, 0.86, 0.51}
\definecolor{chromeyellow}{rgb}{1.0, 0.65, 0.0}
\definecolor{brightube}{rgb}{0.82, 0.62, 0.91}
\definecolor{CornflowerBlue}{RGB}{114, 147, 203}
\newcommand{\cornflowerblue}[1]{\textcolor{CornflowerBlue}{#1}}

\usepackage{interval}
\usepackage{tabularx}
\usepackage{color}
\usepackage{xspace}
\usepackage{mathtools}  % for conditions in math
\usepackage{siunitx}    % for units
\usepackage{placeins}

\interspeechcameraready 
\title{Improving Text-To-Audio Models with Synthetic Captions}

\name[affiliation=1]{Zhifeng}{Kong$^*$}
\name[affiliation=1]{Sang-gil}{Lee$^*$}
\name[affiliation=2]{Deepanway}{Ghosal}
\name[affiliation=2]{Navonil}{Majumder}
\name[affiliation=2]{Ambuj}{Mehrish}
\name[affiliation=1]{\\Rafael}{Valle}
\name[affiliation=2]{Soujanya}{Poria}
\name[affiliation=1]{Bryan}{Catanzaro}

\address{
  $^1$NVIDIA~~~~
  $^2$Singapore University of Technology and Design
}
\email{\{zkong,sanggill,rafaelvalle\}@nvidia.com, sporia@sutd.edu.sg}

\keywords{Text-to-Audio, Text-to-Music, Audio Captioning.}

\begin{document}
\maketitle

\begin{abstract}
It is an open challenge to obtain high quality training data, especially captions, for text-to-audio models. Although prior methods have leveraged \textit{text-only language models} to augment and improve captions, such methods have limitations related to scale and coherence between audio and captions. In this work, we propose an audio captioning pipeline that uses an \textit{audio language model} to synthesize accurate and diverse captions for audio at scale. We leverage this pipeline to produce a dataset of synthetic captions for AudioSet, named \texttt{AF-AudioSet}, and then evaluate the benefit of pre-training text-to-audio models on these synthetic captions. Through systematic evaluations on AudioCaps and MusicCaps, we find leveraging our pipeline and synthetic captions leads to significant improvements on audio generation quality, achieving a new \textit{state-of-the-art}. 
\end{abstract}

\blfootnote{\hspace{-0.5em}$^*$Equal contribution.}
\section{Introduction} \label{sec:introduction}
% text-to-audio
There has been great progress in generative models that generate audio given text descriptions. These models are called text-to-audio (TTA) models \cite{liu2023audioldm, huang2023make, ghosal2023text, kreuk2022audiogen, agostinelli2023musiclm}, and have great potential in a wide range of tasks such as music composition, interactive art, media creation, and education. They also play a critical role in building general purpose multimodal models and agents that can understand and simulate the world in multiple modalities.

% challenges: data; prior 2 ways
Training large-scale text-to-audio models, however, is very challenging. In contrast to the text-to-image domain where there are millions of high-quality samples available \cite{schuhmann2021laion}, there are much fewer high-quality training samples (i.e. audio and caption pairs) in the text-to-audio domain. ~\footnote{Public datasets contain about 0.5K hours of audio and about 100K captions. It is challenging to train large-scale models on these data.}
Meanwhile, the benefits of scaling both compute and data, especially during the pre-training phase, have been emphasized in recent research~\cite{kaplan2020scaling,hoffmann2022training,hagele2024scaling}. In consonance with these findings, this paper shows that pre-training on \textit{high quality} datasets, even if they are synthetic, can drastically improve the quality of text-to-audio models.

In order to create a large dataset for pre-training, prior methods either transform tags and labels into natural language \cite{mei2023wavcaps, melechovsky2023mustango}, or augment audio and captions through mixing and concatenation \cite{huang2023make, ghosal2023text}. These approaches are limited because they require transforming pre-existing metadata, which can be of low quality and result in inconsistencies between the transformed metadata and the audio.

% how we address
In this paper, we propose an alternative approach to obtain high quality audio captions that can be produced at scale and that are based on the audio content. Our approach uses a pre-trained audio language model to automatically caption audio in the wild. Our approach does not require annotation nor metadata associated with audio and, as such, it can be easily scaled-up. 

Automatically captioning audios in the wild, however, has several major challenges. First, the audio language model needs to generalize well to a wide range of audio contents. Second, the generated captions need to be diverse. Finally, given the variability in the quality of generated captions, a mechanism is needed to rank and filter generated captions. We address the first  challenge by adopting the recently proposed Audio Flamingo chat model \cite{kong2024audio} trained on diverse dialogues. We ensure that the captions are diverse by generating captions on the diverse AudioSet dataset \cite{gemmeke2017audio}. Last, to promote the accuracy of the generated captions, we filter them based on their CLAP similarities with the corresponding audios \cite{wu2023large}. With these strategies in place, we are able to generate a large, diverse and high quality dataset of synthetic captions called \texttt{AF-AudioSet}.

% our results
We use text-to-audio (AudioCaps \cite{kim2019audiocaps}) and text-to-music (MusicCaps \cite{agostinelli2023musiclm}) benchmarks to evaluate the benefits of using our method and synthetic captions dataset during pre-training. We systematically study different data filtering and combination strategies, model sizes, as well as commonly used architectural designs based on Tango \cite{ghosal2023text}. We find the optimal pre-training recipes to be consistent across many settings, and with these recipes, we are able to achieve the \textit{state-of-the-art} audio generation quality on both benchmarks. 
To the best of our knowledge, this is the first systematic study to create large-scale high-quality synthetic captions using audio language models and verify their effectiveness in improving text-to-audio models.~\footnote{\scriptsize 
AF-AudioSet: \url{https://github.com/NVIDIA/audio-flamingo/blob/main/labeling_machine}. 
Demos: \url{https://huggingface.co/spaces/declare-lab/Tango-AF}; \url{https://huggingface.co/spaces/declare-lab/Tango-Music-AF}. 
Checkpoints: \url{https://huggingface.co/declare-lab/tango-af-ac-ft-ac}; \url{https://huggingface.co/declare-lab/tango-music-af-ft-mc}.
}

% our contribution
In summary, our contributions are as follows:
\begin{itemize}
    \item We propose a data labeling pipeline to generate large-scale high-quality synthetic captions for audio.
    \item We introduce \texttt{AF-AudioSet}: a large, diverse, and high-quality synthetic caption dataset produced with our pipeline.
    \item We obtain \textit{state-of-the-art} models on text-to-audio and text-to-music through pre-training on \texttt{AF-AudioSet}, and conduct systematic study across a variety of settings.
\end{itemize}
\section{Related work} \label{sec:related_work}
\subsection{Diffusion-based Text-to-Audio Generation}
The research community has made significant progress in diffusion-based \cite{ho2020denoising,rombach2022high} text-to-audio generation models, with recent examples including AudioLDM \cite{liu2023audioldm, liu2024audioldm2}, Make-An-Audio \cite{huang2023make}, and Tango \cite{ghosal2023text, majumder2024tango}. These models use a pre-trained text encoder (e.g., CLAP \cite{elizalde2023clap, wu2023large}, T5 \cite{raffel2020exploring}, or FLAN-T5 \cite{chung2024scaling}) to obtain text embeddings, and a pre-trained variational autoencoder (VAE) \cite{kingma2013auto} to obtain latent features of audio. Similar to latent diffusion models (LDM) \cite{rombach2022high}, the diffusion decoder is trained to generate the audio latent features conditioned on the text embeddings. The generated latent is decoded to a mel spectrogram  representation using the VAE, followed by a neural vocoder \cite{kong2020hifi, lee2022bigvgan} that converts the mel spectrogram into waveform.

\subsection{Training Data Augmentation}
Obtaining diverse, large-scale, and high-quality training data, specially captions, is one of the major challenges in training high-quality text-to-audio models. Especially, a very limited amount of accurate audio-caption pairs are available. Conversely, it is possible to leverage human annotators to produce high quality captions such as AudioCaps \cite{kim2019audiocaps} and MusicCaps \cite{agostinelli2023musiclm}. However, such datasets are very small, e.g. less than 10,000 samples, making it challenging to train large text-to-audio models. 

The current main approach to augment audio captions is to use a large language model to rephrase tags and labels into short captions \cite{mei2023wavcaps}. While this approach can scale-up captions to some extent, it is limited by the existence and quality of the metadata. 
Other approaches focus on augmenting the audio data by concatenating or mixing two samples to form new samples\cite{liu2023audioldm, huang2023make, ghosal2023text}. Though these methods can improve concept-composition capabilities, combining captions is a non-trivial task given that the combination of sounds can result in different captions. \footnote{Imagine captions for the sounds of racing cars and people screaming with and without the sound of gun shots.}

In this paper, we introduce an alternative approach where we label audio based on an audio language model. Our approach can generate high-quality captions as it listens to the audio contents, and can be scaled-up as it does not require any metadata to be provided. 
%It is made possible with recent progress in the audio-language domain \cite{kong2024audio,deshmukh2023pengi,gong2023listen,chu2023qwen,tang2023salmonn}. 
We generate over 600K diverse captions on AudioSet, and find that it can effectively enhance the generation quality of text-to-audio models. To the best of our knowledge, this is the first study that uses an audio language model to create synthetic captions and use them to train and improve text-to-audio models. 
%Though our method could also be used to improve captions generated by combining sounds, we save this analysis for future work. 

\subsection{Audio Captioning Models}
There are several audio-language models that can generate audio captions: Pengi  \cite{deshmukh2023pengi}, LTU \cite{gong2023listen}, Qwen-Audio \cite{chu2023qwen}, Salmonn \cite{tang2023salmonn}, and Audio Flamingo \cite{kong2024audio}. They use different methods to extract audio features and integrate these features into a large language model. 
%Pengi \cite{deshmukh2023pengi} and Audio Flamingo \cite{kong2024audio} use CLAP \cite{elizalde2023clap, elizalde2024natural}, LTU \cite{gong2023listen} and Qwen-Audio \cite{chu2023qwen} use Whisper \cite{radford2023robust}, and SALMONN \cite{tang2023salmonn} combines Whisper and BEATs \cite{chen2023beats}. Pengi, LTU, and Qwen-Audio concatenate audio features with text tokens as inputs to the language model, whereas Audio Flamingo uses cross attention to fuse audio and text features. 
Qwen-Audio and Salmonn are more focused on speech related tasks, while Pengi, LTU, and Audio Flamingo are more focused on non-speech audio understanding. Audio Flamingo in addition provides a chat model trained on diverse dialogues, which can generate more natural and diverse descriptions. Therefore, we use this model in our data synthesis pipeline.

\section{Method and Experimental Setup}\label{sec:methodology}
In this section, we introduce our captioning method and the large-scale synthetic dataset which we call \texttt{AF-AudioSet}. We also introduce our text-to-audio pretraining and finetuning setup. 

\subsection{Generating \texttt{AF-AudioSet}}
We use Audio Flamingo \cite{kong2024audio} to generate captions for audio in the unbalanced training set of AudioSet \cite{gemmeke2017audio}. Audio Flamingo has two series of models. The foundation model is trained on a number of benchmarking datasets including captioning, question-answering, and classification. The chat model is further finetuned on dialogues with more diverse questions and instructions. We investigated both models and found that the synthetic captions from the chat model are more natural and diverse, and therefore decided to use the chat model. Specifically, we prompt the model with the following instruction: \textit{``Can you briefly describe what you hear in this audio?''}. During inference, we generate 20 captions per audio by sampling Audio Flamingo with top-$k=50$ and top-$p=95\%$.

\begin{figure}[!t]
    \centering
    \includegraphics[width=0.45\textwidth]{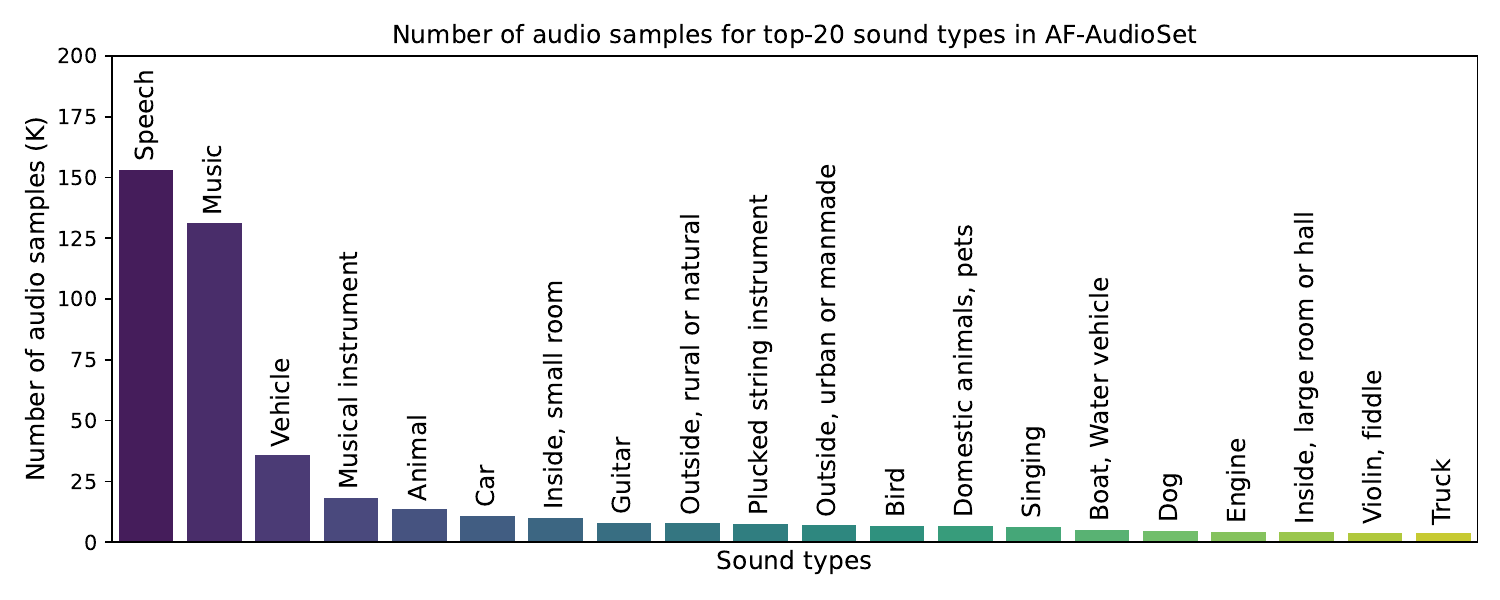}
    \vspace{-1em}
    \caption{Distribution of sound types in \texttt{AF-AudioSet}.}
    \vspace{-1em}
    \label{fig: sound type distribution}
\end{figure}

Given that there is variation in the quality of the generated captions and that we want to promote captions with higher quality, we use the CLAP similarity \cite{wu2023large} between the caption and the audio to rank and filter the synthetic captions. The similarity is computed as $\cos{(\mathbf{v}_{\mathrm{text}}, \mathbf{v}_{\mathrm{audio}})}$, where $\mathbf{v}_{\mathrm{text}}$ is the CLAP text embedding and $\mathbf{v}_{\mathrm{audio}}$ is the CLAP audio embedding. We then store the Top-3 most correlated captions for each audio, and remove captions whose cosine similarities are $<35\%$. We call the filtered synthetic caption dataset \texttt{AF-AudioSet}. In Table \ref{tab: dataset statistics} we demonstrate the number of captions and audio available at different CLAP similarity thresholds. We demonstrate the distribution of sound types in Figure \ref{fig: sound type distribution}.

\begin{table}[!t]
    \centering
    \caption{Number of captions and audio in \texttt{AF-AudioSet} available at different $\tau$, the CLAP similarity threshold.}
    \vspace{-1em}
    \begin{tabular}{c|cccc}
    \toprule
        $\tau$ & $35\%$ & $40\%$ & $45\%$ & $50\%$ \\ \hline
        \# Captions & $696,079$ & $366,018$ & $164,756$ & $61,225$ \\
        \# Audios & $331,421$ & $188,537$ & $91,923$ & $37,220$ \\
    \bottomrule
    \end{tabular}
    \label{tab: dataset statistics}
\end{table}

\subsection{Text-to-Audio Setup}
To systematically study the effect of pretraining text-to-audio models on \texttt{AF-AudioSet}, we use the Tango model \cite{ghosal2023text} with a variety of experimental choices. Tango has three major components: a frozen audio VAE from AudioLDM \cite{liu2023audioldm}, the FLAN-T5 text encoder \cite{chung2024scaling}, and a latent diffusion model that models the latent space of the audio VAE and is conditioned on the text embeddings. The HiFi-GAN vocoder \cite{kong2020hifi} then turns generated mel spectrogram into waveform. 

\textbf{Model size.} We consider three model sizes: small, medium, and large, each with different number of channels. The medium one is the same as \cite{ghosal2023text} with 866M parameters. The large one has 1.93B parameters, and the small one has 217M parameters. Following \cite{liu2023audioldm}, we also investigate replacing the FLAN-T5 text encoder \cite{chung2024scaling} with the CLAP text encoder \cite{wu2023large}, followed by FiLM \cite{perez2018film} conditioning layers -- which we call Tango-CLAP.

\textbf{Pretraining dataset.} We stufy subsets of \texttt{AF-AudioSet} with the four CLAP thresholds $\tau$ shown in Table \ref{tab: dataset statistics}. By changing $\tau$ values we can investigate the trade-off between size and quality in our synthetic data. We also use the same audio captioning pipeline to generate additional captions for AudioCaps -- which can be seen as data augmentation -- and investigate the effect of pretraining on this augmented dataset. Finally, we investigate mixing synthetic and real data during pretraining. 

\textbf{Tasks.} We run experiments on both text-to-audio on AudioCaps \cite{kim2019audiocaps} and text-to-music on MusicCaps \cite{agostinelli2023musiclm}. Specifically, we finetune (pretrained or non-pretrained) Tango models on the train split of either dataset, and run evaluation on their test split. 

\textbf{Metrics.} We report the Frechet Distance (FD) \cite{heusel2017gans}, Frechet Audio Distance (FAD) \cite{roblek2019fr}, Inception Score (IS) \cite{salimans2016improved} with PANNs audio classifier backbone \cite{kong2020panns}, and CLAP similarity \cite{wu2023large} with the 630k-best checkpoint. \footnote{AudioLDM \cite{liu2023audioldm} suggested that FD is preferred over FAD as FD uses a higher quality audio classifier (PANNs) \cite{kong2020panns}.} 

\textbf{Training Setup.} In all experiments, we use 8 A100 GPUs to train the models. We pretrain with a batchsize of 128 for 100K iterations, and finetune with a batchsize of 48 for 40 epochs. The optimization method follows Tango \cite{ghosal2023text}.
\section{Experiments}\label{sec:experiments}
In this section, we aim to answer the following questions: 
\textit{1)} Does pretraining on \texttt{AF-AudioSet} improve generation quality? 
\textit{2)} What is the optimal quality-size trade-off (i.e., $\tau$)? 
\textit{3)} What are the best recipes for different, text encoders, model sizes, and downstream tasks? 
\textit{4)} Does mixing synthetic and real captions during pretraining improve generation quality?

\definecolor{CustomBlue}{RGB}{76, 114, 176}
\definecolor{CustomRed}{RGB}{196, 78, 82}

\begin{figure*}[!t]
    \centering
    \ref{plot: Tango AudioCaps medium: pretrained} Tango (pretrained on \texttt{AF-AudioSet}) \quad
    \ref{plot: Tango AudioCaps medium: not pretrained} Tango (not pretrained)
    
    \begin{minipage}{0.32\textwidth}
    \begin{tikzpicture}
    \begin{axis}[
        xlabel={$\tau$},
        ylabel={FD $\downarrow$},
        xmin=0.35, xmax=0.50,
        ymin=18, ymax=23,
        xtick={0.35,0.40,0.45,0.50},
        ytick={18, 19, 20, 21, 22},
        width=\linewidth, 
        height=3cm
    ]
    \addplot[color=CustomBlue,mark=triangle*,line width=0.8pt] coordinates {(0.35,22.23)(0.40,18.77)(0.45,18.5)(0.50,19.17)};
    \label{plot: Tango AudioCaps medium: pretrained}

    \addplot[color=CustomRed, dashed, line width=0.8pt] coordinates {(0.35,21.71)(0.40,21.71)(0.45,21.71)(0.50,21.71)};
    \label{plot: Tango AudioCaps medium: not pretrained}

    \end{axis}
    \end{tikzpicture}
    \end{minipage}
    \begin{minipage}{0.32\textwidth}
    \begin{tikzpicture}
    \begin{axis}[
        xlabel={$\tau$},
        ylabel={IS $\uparrow$},
        xmin=0.35, xmax=0.50,
        ymin=8.5, ymax=12,
        xtick={0.35,0.40,0.45,0.50},
        ytick={9, 10, 11, 12},
        width=\linewidth,
        height=3cm
    ]

    \addplot[color=CustomBlue,mark=triangle*,line width=0.8pt] coordinates {(0.35,9.81)(0.40,10.42)(0.45,11.11)(0.50,10.94)};

    \addplot[color=CustomRed, dashed, line width=0.8pt] coordinates {(0.35,9.07)(0.40,9.07)(0.45,9.07)(0.50,9.07)};

    \end{axis}
    \end{tikzpicture}
    \end{minipage}
    \begin{minipage}{0.32\textwidth}
    \begin{tikzpicture}
    \begin{axis}[
        xlabel={$\tau$},
        ylabel={CLAP $\uparrow$},
        xmin=0.35, xmax=0.50,
        ymin=0.48, ymax=0.51,
        xtick={0.35,0.40,0.45,0.50},
        ytick={0.48, 0.49, 0.50, 0.51},
        width=\linewidth,
        height=3cm
    ]
    \addplot[color=CustomBlue,mark=triangle*,line width=0.8pt] coordinates {(0.35,0.492)(0.40,0.501)(0.45,0.503)(0.50,0.489)};

    \addplot[color=CustomRed, dashed, line width=0.8pt] coordinates {(0.35,0.50)(0.40,0.50)(0.45,0.50)(0.50,0.50)};

    \end{axis}
    \end{tikzpicture}
    \end{minipage}

    \vspace{-1em}
    \caption{Evaluation results on \textbf{AudioCaps} with different CLAP thresholds of \texttt{AF-AudioSet}. The model is \textbf{Tango (medium)} finetuned on AudioCaps. $\tau=\mathbf{0.45}$ leads to the best results overall and significant improvements over the non-pretrained one.}
    
    \label{fig: Tango AudioCaps medium}
\end{figure*}
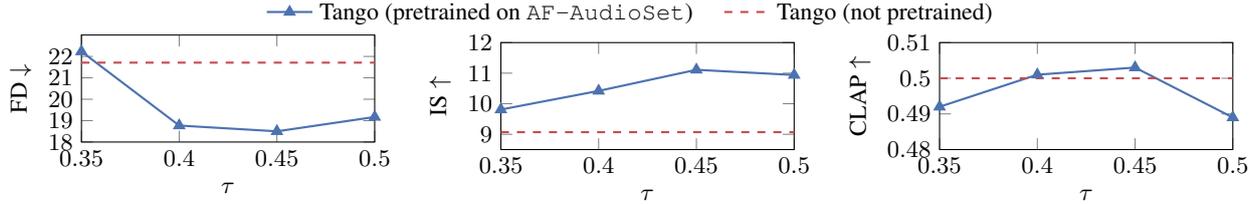

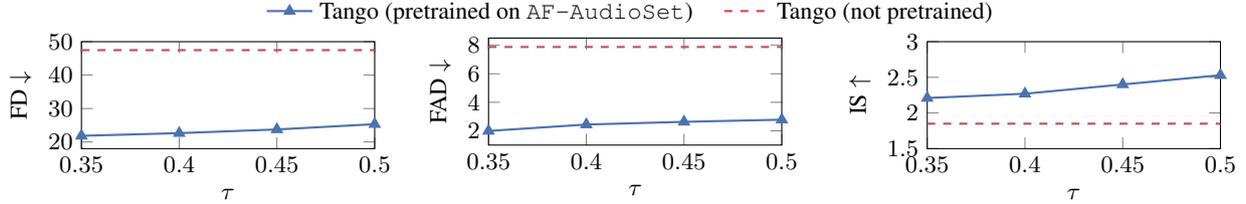
\begin{figure*}[!t]
    \centering
    \ref{plot: Tango MusicCaps medium: pretrained} Tango (pretrained on \texttt{AF-AudioSet}) \quad
    \ref{plot: Tango MusicCaps medium: not pretrained} Tango (not pretrained)
    
    \begin{minipage}{0.32\textwidth}
    \begin{tikzpicture}
    \begin{axis}[
        xlabel={$\tau$},
        ylabel={FD $\downarrow$},
        xmin=0.35, xmax=0.50,
        ymin=18, ymax=50,
        xtick={0.35,0.40,0.45,0.50},
        ytick={20, 30, 40, 50},
        width=\linewidth, 
        height=3cm
    ]
    \addplot[color=CustomBlue,mark=triangle*,line width=0.8pt] coordinates {(0.35,21.84)(0.40,22.64)(0.45,23.74)(0.50,25.32)};
    \label{plot: Tango MusicCaps medium: pretrained}

    \addplot[color=CustomRed, dashed, line width=0.8pt] coordinates {(0.35,47.47)(0.40,47.47)(0.45,47.47)(0.50,47.47)};
    \label{plot: Tango MusicCaps medium: not pretrained}

    \end{axis}
    \end{tikzpicture}
    \end{minipage}
    \begin{minipage}{0.32\textwidth}
    \begin{tikzpicture}
    \begin{axis}[
        xlabel={$\tau$},
        ylabel={FAD $\downarrow$},
        xmin=0.35, xmax=0.50,
        ymin=1, ymax=8.5,
        xtick={0.35,0.40,0.45,0.50},
        ytick={2, 4, 6, 8},
        width=\linewidth,
        height=3cm
    ]
    \addplot[color=CustomBlue,mark=triangle*,line width=0.8pt] coordinates {(0.35,1.99)(0.40,2.44)(0.45,2.63)(0.50,2.78)};

    \addplot[color=CustomRed, dashed, line width=0.8pt] coordinates {(0.35,7.88)(0.40,7.88)(0.45,7.88)(0.50,7.88)};

    \end{axis}
    \end{tikzpicture}
    \end{minipage}
    \begin{minipage}{0.32\textwidth}
    \begin{tikzpicture}
    \begin{axis}[
        xlabel={$\tau$},
        ylabel={IS $\uparrow$},
        xmin=0.35, xmax=0.50,
        ymin=1.5, ymax=3,
        xtick={0.35,0.40,0.45,0.50},
        ytick={1.5, 2, 2.5, 3},
        width=\linewidth,
        height=3cm
    ]
    \addplot[color=CustomBlue,mark=triangle*,line width=0.8pt] coordinates {(0.35,2.21)(0.40,2.27)(0.45,2.40)(0.50,2.53)};

    \addplot[color=CustomRed, dashed, line width=0.8pt] coordinates {(0.35,1.85)(0.40,1.85)(0.45,1.85)(0.50,1.85)};

    \end{axis}
    \end{tikzpicture}
    \end{minipage}%
    
    \vspace{-1em}
    \caption{Evaluation results on \textbf{MusicCaps} with different CLAP thresholds of \texttt{AF-AudioSet}. The model is \textbf{Tango (medium)} finetuned on MusicCaps. $\tau=\mathbf{0.35}$ leads to the best FD and FAD and significant improvements over the non-pretrained one.}
    \vspace{-0.5em}
    
    \label{fig: Tango MusicCaps medium}
\end{figure*}

\subsection{Pretraining Leads to SOTA Text-to-Audio Quality}
\label{sec: experiment sota}

Our text-to-audio results on AudioCaps are shown in Table \ref{tab: audiocaps main}. Tango-FT-AC refers to the \textit{baseline} Tango without pretraining, and \cornflowerblue{Tango-Full}-FT-AC refers to the one pretrained on TangoPromptBank \cite{ghosal2023text}. \orange{Tango-AF}-FT-AC refers to Tango pre-trained on \texttt{AF-AudioSet} ($\tau=0.45$), and \orange{Tango-AF\&AC}-FT-AC refers to Tango pre-trained on \texttt{AF-AudioSet} ($\tau=0.45$) + AudioCaps. After pretraining, these models are finetuned on AudioCaps. We find that pretraining on \texttt{AF-AudioSet} leads to a systematic improvement over the Tango baseline, especially in IS. We also find that pretraining on \texttt{AF-AudioSet} + AudioCaps results in further improvements, especially in FD. Positively, our best results also outperform recent state-of-the-art results. 

The text-to-music results on MusicCaps are shown in Table \ref{tab: musiccaps main}. Tango-FT-MC refers to the \textit{baseline} Tango without pretraining, and \cornflowerblue{Tango-Full}-FT-MC refers to the one pretrained on TangoPromptBank \cite{ghosal2023text}. \orange{TangoMusic-AF}-FT-MC refers to Tango pre-trained on \texttt{AF-AudioSet} ($\tau=0.35$). All models are then finetuned on MusicCaps. After pretraining on \texttt{AF-AudioSet}, the model significantly improves on all metrics and outperforms recent state-of-the-art baselines.

We summarize the results as the major finding of this paper:
\begin{tcolorbox}[boxsep=1pt,left=2pt,right=2pt,top=1pt,bottom=1pt,colback=blue!5!white,colframe=gray!75!black]
Pretraining Tango on \texttt{AF-AudioSet} can lead to state-of-the-art text-to-audio and text-to-music generation quality.
\end{tcolorbox}

\begin{table}[!t]
    \centering
    \caption{Evaluation results on AudioCaps. Pretraining on \texttt{AF-AudioSet} (\orange{Tango-AF}-FT-AC) leads to consistent improvement over the non-pretrained one (Tango-FT-AC). Pretraining on a mix of \texttt{AF-AudioSet} and AudioCaps (\orange{Tango-AF\&AC}-FT-AC) further improves and leads to SOTA text-to-audio generation quality. $^{\dag}$ indicates the numbers are taken from their original papers. $^{\ddag}$ indicates the numbers are taken from \cite{majumder2024tango}.}
    \vspace{-1em}
    \begin{tabular}{l|ccc}
    \toprule
        Model & FD $\downarrow$ & CLAP $\uparrow$ & IS $\uparrow$\\ \hline
        AudioLDM-L-Full \cite{liu2023audioldm} & $23.31^{\dag}$ & - & - \\
        Make-an-Audio \cite{huang2023make} & $\underline{18.32}^{\dag}$ & $0.454$ & $7.29^{\dag}$ \\
        CoDi \cite{tang2024any} & $22.90^{\dag}$ & - & $8.77^{\dag}$ \\ 
        ConsistencyTTA \cite{bai2023accelerating} & $20.97^{\dag}$ & $0.496$ & $8.88^{\dag}$ \\
        Auffusion \cite{xue2024auffusion} & $21.99^{\dag}$ & $\mathbf{0.539}$ & $10.57^{\dag}$ \\ \hline
        Tango-FT-AC \cite{ghosal2023text} & $19.84$ & $0.500$ & $9.06$ \\
        \cornflowerblue{Tango-Full}-FT-AC \cite{ghosal2023text} & $18.93^{\dag}$ & $\mathbf{0.539}^{\ddag}$ & $7.86^{\ddag}$ \\
        \orange{Tango-AF}-FT-AC & $19.06$ & $0.503$ & $\underline{10.87}$ \\
        \orange{Tango-AF\&AC}-FT-AC & $\mathbf{17.19}$ & $\underline{0.527}$ & $\mathbf{11.04}$ \\
    \bottomrule
    \end{tabular}
    \label{tab: audiocaps main}
\end{table}

\begin{table}[!t]
    \centering
    \caption{Evaluation results on MusicCaps. Pretraining on \texttt{AF-AudioSet} (\orange{TangoMusic-AF}-FT-MC) significantly outperforms the non-pretrained one (Tango-FT-MC) and leads to SOTA text-to-music generation quality.}
    \vspace{-1em}
    \begin{tabular}{l|ccc}
    \toprule
        Model & FD $\downarrow$ & FAD $\downarrow$ & IS $\uparrow$ \\ \hline
        MusicGen (medium) \cite{copet2024simple} & $35.52$ & $5.02$ & $1.94$ \\
        AudioLDM-2 \cite{liu2024audioldm2} & $22.08$ & $3.83$ & $2.17$ \\ \hline
        Tango-FT-MC & $47.47$ & $7.88$ & $1.85$ \\
        \cornflowerblue{Tango-Full}-FT-MC & $38.19$ & $6.83$ & $\mathbf{2.71}$ \\
        \orange{TangoMusic-AF}-FT-MC & $\mathbf{21.84}$ & $\mathbf{1.99}$ & $2.21$ \\
    \bottomrule
    \end{tabular}
    \vspace{-1em}
    \label{tab: musiccaps main}
\end{table}

\subsection{Trade-off between Synthetic Captions Quality and Size}
\label{sec: experiment tradeoff}

The CLAP threshold $\tau$ controls the trade-off between caption quality and data size in \texttt{AF-AudioSet}. A larger $\tau$ leads to a smaller subset (see Table \ref{tab: dataset statistics}) but the remaining captions are more correlated to the audio given CLAP as a similarity score. In Figures \ref{fig: Tango AudioCaps medium} and \ref{fig: Tango MusicCaps medium}, we plot the evaluation metrics on AudioCaps and MusicCaps with different $\tau$ and using the medium-sized Tango. 
On AudioCaps, $\tau=0.45$ leads to the best results and significantly outperforms the baseline Tango without pretraining. The results monotonically improve as $\tau$ increases from $0.35$ to $0.45$, However, there is result degradation when $\tau$ changes from $0.45$ to $0.5$, indicating the subset with $\tau=0.5$ may be too small for pretraining. 
On MusicCaps, as $\tau$ increases, FD and FAD become slightly worse while IS becomes slightly better. However, the differences are small, and all results are significantly better than the baseline Tango model without pretraining. We think that $\tau=0.35$ leads to the best generation quality as it has the best FD and FAD.
We summarize the recipes below:
\begin{tcolorbox}[boxsep=1pt,left=2pt,right=2pt,top=1pt,bottom=1pt,colback=blue!5!white,colframe=gray!75!black]
$\tau=0.45$ leads to the best results on AudioCaps.\\
$\tau=0.35$ leads to the best results on MusicCaps.
\end{tcolorbox}

\subsection{\texttt{AF-AudioSet} versus TangoPromptBank}

We also investigate pretraining on TangoPromptBank \cite{ghosal2023text} -- a collected pretraining set with more than 1M samples -- and fintune on both benchmarks. The resulting models are called \cornflowerblue{Tango-Full}-FT-AC(or MC), and results are in Tables \ref{tab: audiocaps main} and \ref{tab: musiccaps main}. We find pretraining on \texttt{AF-AudioSet} (\orange{Tango-AF}-$*$) can match the generation quality as pretraining on TangoPromptBank on AudioCaps and significantly outperforms it on MusicCaps. The results indicate that \texttt{AF-AudioSet} has high quality captions, and pretraining on this smaller yet higher quality set can lead to similar or better results. We summarize our findings below:
\begin{tcolorbox}[boxsep=1pt,left=2pt,right=2pt,top=1pt,bottom=1pt,colback=blue!5!white,colframe=gray!75!black]
Pretraining on high-quality datasets, even if they are synthetic and smaller, can lead to similar or better generation quality.
\end{tcolorbox}

\subsection{The Effect of Text Encoder and Tango Size}
\label{sec: experiment ablation}
We investigate the optimal filtering threshold $\tau$ for Tango-CLAP, where we replace the FLAN-T5 text encoder with CLAP. The results in Figure \ref{fig: Tango_CLAP AudioCaps medium} are very similar to Tango and the best results occur at $\tau=0.45$.

We then study the effect of our synthetic data on different Tango model sizes: small, medium, and large, and compare the results with and without pretraining. Figures \ref{fig: Tango AudioCaps size} and \ref{fig: Tango MusicCaps size} show results on AudioCaps and MusicCaps. For all  sizes, pretraining leads to significant improvements, indicating that pretraining on \texttt{AF-AudioSet} is a efficient and versatile strategy to improve audio generation quality. 
We summarize our the findings below:
\begin{tcolorbox}[boxsep=1pt,left=2pt,right=2pt,top=1pt,bottom=1pt,colback=blue!5!white,colframe=gray!75!black]
The recipes and conclusions in Section \ref{sec: experiment tradeoff} also apply to other model conditioning architectures and model sizes.
\end{tcolorbox}

\begin{figure}[!t]
    \centering
    \begin{tikzpicture}
    \begin{axis}[
        xlabel={$\tau$},
        ylabel={FD $\downarrow$},
        xmin=0.35, xmax=0.50,
        ymin=24, ymax=35,
        xtick={0.35,0.40,0.45,0.50},
        ytick={25, 30, 35},
        width=\linewidth, 
        height=4cm
    ]
    \addplot[color=CustomBlue,mark=triangle*,line width=0.8pt] coordinates {(0.35,30.51)(0.40,26.37)(0.45,24.89)(0.50,28.15)};
    \addlegendentry{\footnotesize{Tango-CLAP (pretrained on \texttt{AF-AudioSet})}};

    \addplot[color=CustomRed, dashed, line width=0.8pt] coordinates {(0.35,26.86)(0.40,26.86)(0.45,26.86)(0.50,26.86)};
    \addlegendentry{\footnotesize{Tango-CLAP (not pretrained)}};

    \end{axis}
    \end{tikzpicture}

    \vspace{-1em}
    \caption{Evaluation results on \textbf{AudioCaps} with different CLAP thresholds of \texttt{AF-AudioSet}. The model is \textbf{Tango-CLAP (medium)} finetuned on AudioCaps. The results are similar to Tango in Figure \ref{fig: Tango AudioCaps medium}.}
    \vspace{-1em}
    
    \label{fig: Tango_CLAP AudioCaps medium}
\end{figure}
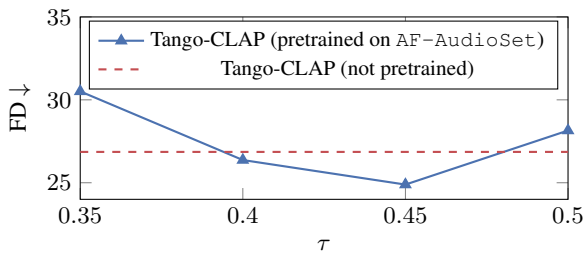

\begin{figure}[!t]
    \centering
    \begin{minipage}{0.46\textwidth}
    \begin{tikzpicture}
    \begin{axis}[
        xlabel={$\tau$},
        ylabel={FD $\downarrow$},
        xmin=1, xmax=3,
        ymin=18, ymax=28,
        xtick={1, 2, 3},
        xticklabels={small, medium, large},
        ytick={20,23,26},
        width=\linewidth, 
        height=4cm
    ]
    \addplot[color=CustomBlue,mark=triangle*,line width=0.8pt] coordinates {(1,20.46)(2,18.50)(3,19.06)};
    \addlegendentry{\footnotesize{Tango (pretrained on \texttt{AF-AudioSet})}};

    \addplot[color=CustomRed,mark=triangle*,line width=0.8pt] coordinates {(1,25.62)(2,21.71)(3,19.84)};
    \addlegendentry{\footnotesize{Tango (not pretrained)}};

    \end{axis}
    \end{tikzpicture}

    \vspace{-1em}
    \caption{Evaluation results on \textbf{AudioCaps} with different model sizes. The model is \textbf{Tango} pre-trained on \texttt{AF-AudioSet} with $\tau=0.45$ and finetuned on AudioCaps. The improvement by pretraining is clear across all model sizes.}
    \vspace{1em}
    \label{fig: Tango AudioCaps size}
    \end{minipage}
    \begin{minipage}{0.46\textwidth}
    \centering
    \begin{tikzpicture}
    \begin{axis}[
        xlabel={$\tau$},
        ylabel={FD $\downarrow$},
        xmin=1, xmax=3,
        ymin=18, ymax=80,
        xtick={1, 2, 3},
        xticklabels={small, medium, large},
        ytick={20,40,60},
        width=\linewidth, 
        height=4cm
    ]
    \addplot[color=CustomBlue,mark=triangle*,line width=0.8pt] coordinates {(1,25.34)(2,21.84)(3,23.39)};
    \addlegendentry{\footnotesize{Tango (pretrained on \texttt{AF-AudioSet})}};

    \addplot[color=CustomRed,mark=triangle*,line width=0.8pt] coordinates {(1,60.39)(2,47.47)(3,46.93)};
    \addlegendentry{\footnotesize{Tango (not pretrained)}};

    \end{axis}
    \end{tikzpicture}

    \vspace{-1em}
    \caption{Evaluation results on \textbf{MusicCaps} with different model sizes. The model is \textbf{Tango} pre-trained on \texttt{AF-AudioSet} with $\tau=0.35$ and finetuned on MusicCaps. The improvement by pretraining is clear across all model sizes.}
    \vspace{-1em}
    \label{fig: Tango MusicCaps size}
    \end{minipage}
    
\end{figure}

\subsection{The Effect of Mixed pretraining Sets}
\label{sec: experiment mix pretraining}

Finally, we study the effect of several mixed pretraining sets, where we combine synthetic and real captions during pretraining. First, we look at combining \texttt{AF-AudioSet} and AudioCaps. The results are in Table \ref{tab: audiocaps main}, with model name \orange{Tango-AF\&AC}-FT-AC. The results show that combining both datasets during pretraining improves generation quality. Then, we look at augmenting AudioCaps with synthetic captions generated with our pipeline described in Section \ref{sec:methodology}. The results for this setting are: FD$=19.59$, CLAP$=0.507$, and IS$=9.85$. The results  show that simply augmenting AudioCaps leads to consistently better generation quality. We summarize our findings below:
\begin{tcolorbox}[boxsep=1pt,left=2pt,right=2pt,top=1pt,bottom=1pt,colback=blue!5!white,colframe=gray!75!black]
Combining synthetic and real data during pretraining can lead to further improvements on generation quality.
\end{tcolorbox}

\section{Discussion}\label{sec:discussion}

We expect that the quality of synthetic captions will improve, as audio language models become larger and more powerful -- including audio captioning models and contrastive audio-text embeddings. An important future direction is to investigate a better synthesis pipeline to further improve diversity and accuracy of synthetic captions. Another important future direction is to investigate better pretraining strategies.

% \clearpage
% \newpage
\bibliographystyle{IEEEtran}
\bibliography{main}
\end{document}